\documentclass[runningheads]{llncs}
\usepackage{graphicx}
\usepackage{booktabs}
\usepackage{multirow}
\usepackage{array}
\usepackage{adjustbox}
\usepackage[accsupp]{axessibility}

\usepackage{amsmath}
\usepackage{fancyhdr}

\usepackage{orcidlink}
\title{Key-point Guided Deformable Image Manipulation Using Diffusion Model}
\author{Seok-Hwan Oh*$^{1}$, Guil Jung*$^{1}$, Myeong-Gee Kim$^{2}$, Sang-Yun Kim$^{1}$, \\
Young-Min Kim$^{1}$, Hyeon-Jik Lee$^{1}$, Hyuk-Sool Kwon$^{3}$, Hyeon-Min Bae$^{1}$}

\institute{
$^{1}$Department of Electrical Engineering, KAIST, Daejeon, South Korea \and
$^{2}$Barreleye Inc., Seoul, South Korea \and
$^{3}$Department of Emergency Medicine, SNUBH, Seong-nam, South Korea
\footnotetext[1]{These authors contributed equally to this work.}}

\begin{document}

\maketitle

\begin{abstract}
  In this paper, we introduce a Key-point guided Diffusion probabilistic Model (KDM) that gains precise control over images by manipulating the object's key-points. We propose a two-stage generative model incorporating an optical flow map as an intermediate output. By doing so, a dense pixel-wise understanding of the semantic relation between the image and sparse key-point is configured, leading to more realistic image generation. Additionally, the integration of optical flow helps regulate the inter-frame variance of sequential images, demonstrating an authentic sequential image generation. The KDM is evaluated with diverse key-point conditioned image synthesis tasks, including facial image generation, human pose synthesis, and echocardiography video prediction, demonstrating the KDM is proving consistency enhanced and photorealistic images compared with state-of-the-art models.
  \keywords{Image manipulation \and diffusion probabilistic model \and video generation}
\end{abstract}

\section{Introduction}
\label{sec:intro}
Deep generative models such as generative adversarial networks (GAN)~\cite{goodfellow2014generative} and denoising diffusion model~\cite{ho2020denoising}, have shown great success in generating realistic images. However, human imagination is not only limited to envisioning virtual objects and scenes, but they are also able to anticipate the movement or change of the entities. Recently, there has been a surge of academic research in converting one image into another, tailored to the user’s specific objective~\cite{pang2021image}. Those image-synthesizing models are applied to diverse areas such as human-computer interaction, virtual reality, and video generation. Furthermore, the generated image can be employed for diverse downstream tasks such as human pose estimation and face landmark detection.

Generating a novel image in accordance with user-specific conditions from an original image presents a significant challenge. The difficulty arises from the fact that such an image synthesis task demands a high-level semantic understanding between the input image and the user's controls. To address the limitations, several GANs are proposed to generate images based on object key-points~\cite{di2018gp,siarohin2018deformable,ma2017pose,tang2019cycle}. However, such GAN-based approaches suffer from unstable training and limited diversity of generated samples~\cite{wiatrak2019stabilizing}. Variational autoencoder~\cite{kingma2013auto} is proposed as an alternative generative model offering relatively stable learning objective, however offers images of limited quality.

More recently, the denoising diffusion probabilistic model (DDPM) has been proposed to synthesize high-quality images through iterative denoising-based image generation~\cite{ho2020denoising}. The DDPM demonstrates the capacity for diverse and high-quality image generation while providing a stable training framework. Ankan et al. adopted the DDPM scheme in order to generate key-point conditioned human pose image synthesis and succeeded in generating realistic images with a semantic understanding of the relation between the key-point and input image~\cite{bhunia2023person}. However, such key-point-driven direct image synthesizing techniques can lead to inconsistencies in sequential image generation. Kim et al. demonstrated that the incorporation of a deformation matrix to DDPM facilitates realistic sequential image generation~\cite{kim2022diffusemorph}. However, relying solely on spatial deformation for image synthesis exhibits limitations in depicting high-frequency details.

To address the challenges, we propose a two-stage diffusion model that utilizes optical flow as an intermediate output, enabling the generation of sequentially consistent images while preserving the authenticity of the input images. In the initial stage, the diffusion model synthesizes the optical flow image by interpreting the semantic relationship between key-point controls and the input images. Subsequently, the second stage diffusion model leverages the synthesized optical flow to generate images reflecting the user control. By employing the second stage diffusion model, we achieve fine-grained image generation preserving details of the original image, surpassing the limitations of direct linear image deformation with optical flow.

The proposed scheme is a cross-modal framework, that employs input image and key-point in an interactive manner. Extensive experiments across three distinct image manipulation tasks are introduced. A facial landmark-based image synthesis and key-point guided human pose generation are experimented and demonstrate that the proposed framework offers sequentially stable and realistic image generation. Additionally, we apply the neural network to echocardiography video generation, proving the framework is capable of generating sophisticated medical videos.

\section{Related work}

\subsection{Conditional latent diffusion model}
DDPM has been proposed to generate realistic images through iterative denoising steps starting from noise following a normal distribution. DDPM is capable of generating high-fidelity images, but it encounters a significant computation during the image sampling process. To address the problem, Yang et al. proposed DDIM~\cite{song2020denoising}, which leverages a non-Markovian process to achieve satisfactory outcomes with a reduced number of denoising steps. Furthermore, to alleviate the computational resources required for pixel-level training and sampling, the latent diffusion model (LDM)~\cite{rombach2022high} was introduced recently. Diverse attempts have been made to provide conditional image generation through diffusion models. The classifier guidance~\cite{dhariwal2021diffusion} method effectively reduces the diversity of conditional diffusion models by utilizing the gradient of pre-trained models. An alternative approach, Classifier-free guidance~\cite{ho2022classifier}, was introduced, eliminating the necessity for classifier training for the conditional denoising process. Such conditional guidance methods have enabled image generation across a wide range of conditions, including both textual and visual resources~\cite{nichol2021glide,saharia2022palette,yang2023paint,hertz2022prompt,couairon2022diffedit}.

\subsection{Unsupervised optical flow estimation}
The term ‘Optical flow’ is used to describe the dense mapping of image correspondence between consecutive frames in video. FlowNet~\cite{dosovitskiy2015flownet} proposed an end-to-end convolutional optical flow network, demonstrating remarkable performance in optical flow estimation. Subsequent developments in network structures and optimization techniques~\cite{ilg2017flownet,sun2018pwc,teed2020raft,sun2019models,yang2019volumetric} have enabled learning-based methods to significantly outperform traditional approaches~\cite{black1993framework,bruhn2005lucas,horn1981determining}. However, such methods rely on supervised learning, which requires an extensive and expansive collection of ground-truth data. To address the challenges, unsupervised optical flow methods have been proposed. The objective function of unsupervised learning minimizes the difference between the original image and the warped image using predicted flow vectors, eliminating the need for labels. Diverse approaches were suggested to enhance precision in unsupervised optical flow estimation, including occlusions addressing~\cite{wang2018occlusion}, depth constraint incorporation~\cite{ranjan2019competitive,zou2018df}, and data distillation employment~\cite{liu2019ddflow}. Notably, ARFlow~\cite{liu2020learning} enhances reliability in unsupervised learning by using augmented images for additional supervision, resulting in improved compatibility and generalization.

\subsection{Key-point guided image-to-image translation}
Key-points represent meaningful locations in images that contain spatial information regarding an object's shape and position. The number and positioning of key-points may vary for the same object, depending on the criteria or detection model in use. Typically, in the context of facial data, \textit{OpenFace}~\cite{baltrusaitis2018openface} and \textit{dlib}~\cite{king2009dlib} propose to use 68 facial landmarks, while for human pose analysis, \textit{OpenPose} suggests using either 18 or 25 key-points~\cite{cao2017realtime}. Since key-points represent the primary features of an object, there is a growing academic interest in facilitating user modifications in images through the adjustment of key-points. 

The key-point guided image-to-image translation represents a methodology for the transformation of source images into desired representations. The task poses a challenge, as it necessitates the generation of photorealistic images while preserving the inherent attributes of the source image. The realm of key-point guided image-to-image translation has experienced substantial performance enhancements, largely attributed to GAN models. Ma et al.~\cite{ma2017pose} pioneered the exploration of key-point-based framework, human pose transfer, employing a dual-generator.  Wang et al.~\cite{wang2018every} proposed a CMM-net applied for face image translation through facial landmark guidance. Subsequently, Tang et al.~\cite{tang2019cycle} introduced a C2GAN scheme implemented based on the CycleGAN~\cite{zhu2017unpaired} methodology, demonstrating an advanced interpretation of the interaction between the image and key-point guidance.
\begin{figure}[t]
\centering
\includegraphics[width=\textwidth]{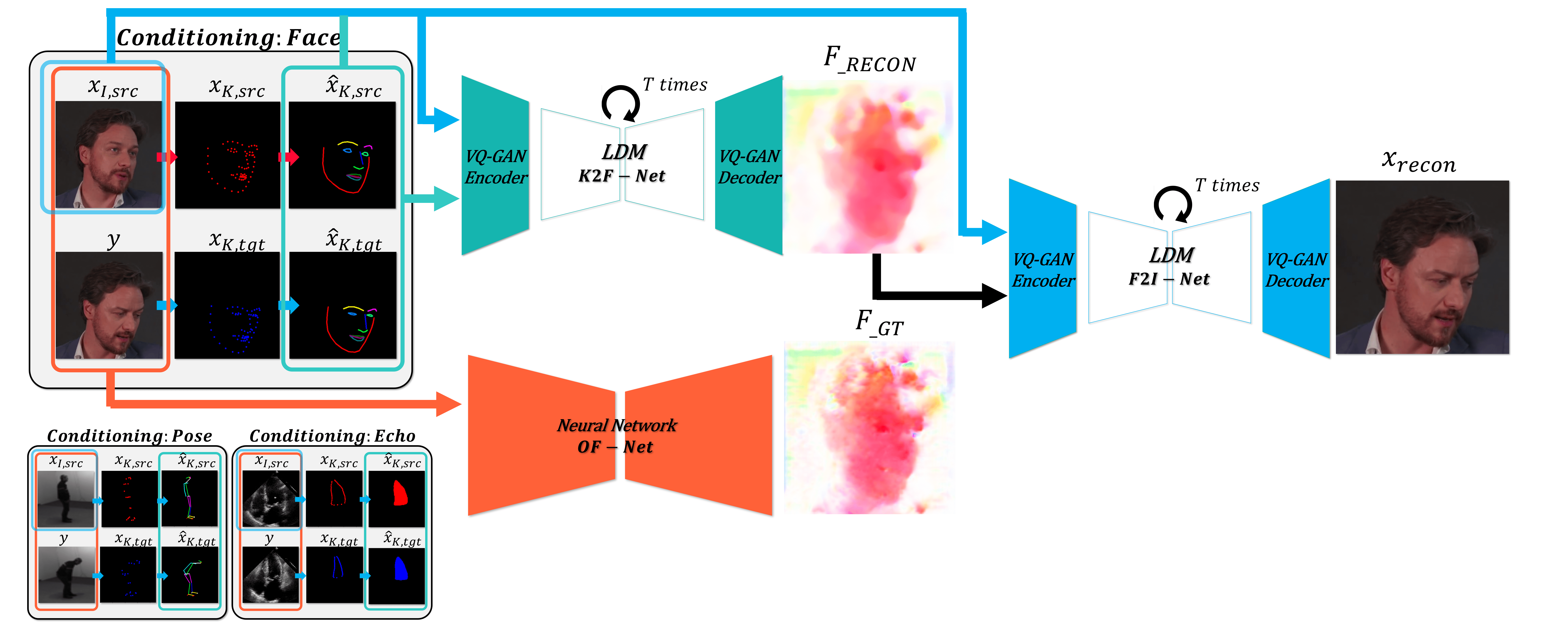}
\caption{Illustration of the proposed KDM framework is presented. The source image and corresponding key-point of the facial expression generation, human pose synthesis, and echocardiography video synthesis are introduced.}
\label{fig1}
\end{figure}

\section{Method}
In this paper, we propose a conditional diffusion-based probabilistic generative model  $p(y|x_{I,src}, x_{K,src}, x_{K,tgt})$ designed to produce an output image that adjusts to the user’s key-point control, while preserving the style of the input image, $x_{I,src}$. Given the conditional generative model, a user can easily generate the desired image only by manipulating the source key-point $x_{K,src}$ to target key-point $x_{K,tgt}$. When it comes to sequential image generation where the users perform multiple key-point editing processes, the existing key-point-based image synthesizing methods \cite{bhunia2023person} encounter difficulties in generating sequentially consistent results, leading to an unnatural user experience.

To address the limitation, we propose to split the generative model
\begin{equation}
\begin{gathered}
    \label{eq:equation1}
    p(y \mid x_{\text{I,src}}, x_{\text{K,src}}, x_{\text{K,tgt}}) \\
    \rightarrow p(y \mid F, x_{\text{I,src}}, x_{\text{K,src}}, x_{\text{K,tgt}}) p(F \mid x_{\text{I,src}}, x_{\text{K,src}}, x_{\text{K,tgt}})
\end{gathered}
\end{equation}
to two sequential procedures of 1. generating optical flow $F$ from the source image and key-point control, and 2. synthesizing output image employing optical flow as input condition. Assisted by the optical flow generation process, the neural network can better interpret the semantic relationship between the key-point and the provided image. In addition, by formulating dense optical flow from sparse key-point condition variables, the uncertainty in the generated image is mitigated, stabilizing sequential image synthesis. Specifically, in the case of single-step key-point guided image synthesis, the background is not regularized by the sparse key-point. As a result, in sequential image generation, maintaining the consistency of each frame image becomes challenging with the single-step model (Figure~\ref{fig3} 11th column demonstrates the inconsistency of the single-step sequential image synthesis). However, the dense optical flow of the Key-point guided Diffusion probabilistic Model (KDM) supervises the uncertainty of the background and helps to provide consistent sequential image generation.

Figure~\ref{fig1} illustrates the overall configuration of the KDM framework. The key-point to the optical flow generation network, K2F-net, and optical flow to image generation network, F2I-net, are implemented based on the denoising diffusion probabilistic model, which is recently proven to provide high-quality image outcome through an iterative denoising process. An optical flow estimation network (OF-net) is proposed to generate optical flow from the $x_{\text{I,src}}$ and $y$. The OF-net provides the optical flow matrix $F_{\text{GT}}$ to K2F-net, so that the K2F-net employs the $F_{\text{GT}}$ as a ground-truth optical flow for training.

We provide a detailed composition of the optical flow estimation module in section~\ref{section31}, and the analysis of key-point to optical flow generation and optical flow to the image synthesizing diffusion model are described in section~\ref{section32} and~\ref{section33}, respectively.

\subsection{Optical flow estimation}
\label{section31}
Given a pair of source $x_{\text{I,src}}$ and target images $y$, the OF-net inferences optical flow matrix, representing per-pixel displacement between the images. The OF-net architecture is modeled based on PWC-net~\cite{sun2018pwc}, which is one of the most widely used baseline optical flow estimation networks due to its efficiency in computation. For the wide-range application of the proposed framework, particularly in scenarios where ground-truth optical flow is unavailable, the OF-net is trained in an unsupervised manner. The OF-net is trained to minimize photo-metric loss between the target image $y$ and $x_{\text{I,src}}(I+F)$ warped source image using the optical flow prediction, where $I$ refers to the identity matrix. The details of the learning scheme are implemented based on~\cite{liu2020learning}.

For ease of training, the OF-net is pre-trained on KITTI dataset \cite{geiger2012we}, and is then fine-tuned with each application dataset. After the optimization of the OF-net, the OF-net is employed for the training of K2F-net. During the training of K2F-net, the OF-net inferences optical flow estimation of $x_{\text{I,src}}$ and $y$, and provide it as the ground-truth of the K2F-net.

\subsection{Key-point guided optical flow synthesis}
\label{section32}
The K2F-net is implemented to generate optical flow from the key-point $x_K$ and the input image $x_{\text{I,src}}$. The key-point was extracted from the $x_{\text{I,src}}$ through widely adopted pre-trained models~\cite{king2009dlib,cao2017realtime}. The key-points are represented in the form of $x_K \sim \mathbb{R}^{N \times 2}$, indicating a 2D pixel position. The number of key-points, denoted by $N$, varies depending on the application, and the order of key-points contains distinct semantic features of the object. Further descriptions of the key-point attributes and corresponding detection models are provided in section~\ref{section41}.

The key-point $x_{K}$ vector can be represented as a spatial domain image $\hat{x}_{K}$. For example, a skeleton image is obtained by connecting adjacent body parts of the pose key-points. The K2F-net is either capable of utilizing $x_{K}$ or $\hat{x}_{K}$ as conditioning inputs. We have empirically observed that the K2F-net better interprets spatial relation between the key-point condition and input images when the key-point condition is provided in spatial format $\hat{x}_{K}$.

Generating a dense pixel-wise deformation matrix from sparse key-point conditions is inherently a \textit{one-to-many problem}, that has multiple plausible outputs for a single pair of inputs. In order to address the problem, the K2F-net adopts a denoising diffusion probabilistic model that represents a feature in a stochastic manner. However, the denoising diffusion model is computationally demanding since the neural network requires several dozens to a thousand inferences for a single image generation. To enhance the computation efficiency of DDPM, the LDM proposes to apply a denoising process in the latent space domain instead of the pixel space domain~\cite{rombach2022high}. As introduced in~\cite{rombach2022high}, the K2F employs VQ-GAN to encode $224 \times 224$ pixel image to $56 \times 56$ latent feature. The iterative denoising procedure is performed in compressed latent space. The high-quality image is then generated by decoding the latent space to pixel space through the VQ-GAN decoder. To match the size of each $\hat{x}_{K}$ to that of the latent feature, a compact condition encoder $\Psi$ is implemented and converts $224 \times 224$ size $\hat{x}_{K}$ to $56 \times 56$ feature. The $\Psi$ is composed of three convolution layers with kernel size $3 \times 3$ and stride $2 \times 2$.

The learning objective of K2F-net is described as follows:
\begin{equation}
L := \mathbb{E}_{\phi_{K2F}, \epsilon \sim \mathcal{N}(0,1)} \left[ \left| \epsilon - \phi_{K2F}(x_{\text{I,src}}, x_{\text{K,src}}, x_{\text{K,tgt}}, t) \right| \right]
\end{equation}

where, $t = (1, 2 , \ldots, T)$ denotes the time steps of the denoising process. The backbone of $\phi_{K2F}$ is time-conditional U-net.

\subsection{Optical flow-guided image synthesis}
\label{section33}
Optical flow, generated from K2F-net, provides per-pixel displacement between $x_{\text{I,src}}$ and $y$. In general, the $x_{\text{I,src}}$ can be warped to produce a deformed image $x_{\text{I,src}}(I + F)$ with the deformation prediction $F$. However, such a linear deformation-based image manipulation scheme demonstrates limitations in describing high-frequency details of the given input image (see Figure~\ref{fig3}). To incorporate input image details into the generation of $x_{\text{recon}}$, we propose to employ a generative model, F2I-net, instead of conventional warping.

The F2I-net utilizes $x_{\text{I,src}}$, $x_K$, and $F$ as inputs for generating a detail-preserved output $y$. In order to generate a high-quality image with reasonable computational requirements, the F2I-net adopts the latent diffusion model as a baseline structure. Similar to the K2F-net, the VQ-GAN is employed for the latent domain representation of the image, and a time-conditional U-net is employed for the denoising process. The conditioning image $x_{\text{I,src}}$, $x_K$, and $F$ are channel-wise concatenated and encoded to the latent-domain feature dimension through a series of $2 \times 2$ strided convolutions. The conditional iterative denoising process synthesizes a latent feature, and through a single decoding procedure of the latent feature, the output image is generated. The training objective of the F2I-net is formulated as
\begin{equation}
L := \mathbb{E}_{\phi_{F2I}, \epsilon \sim \mathcal{N}(0,1)} \left[ \left| \epsilon - \phi_{F2I}(x_{\text{I,src}}, x_K, F, t) \right| \right].
\end{equation}
\subsection{Inference: Initialization with linear deformation}
\begin{figure}[h]
\centering
\includegraphics[width=\textwidth]{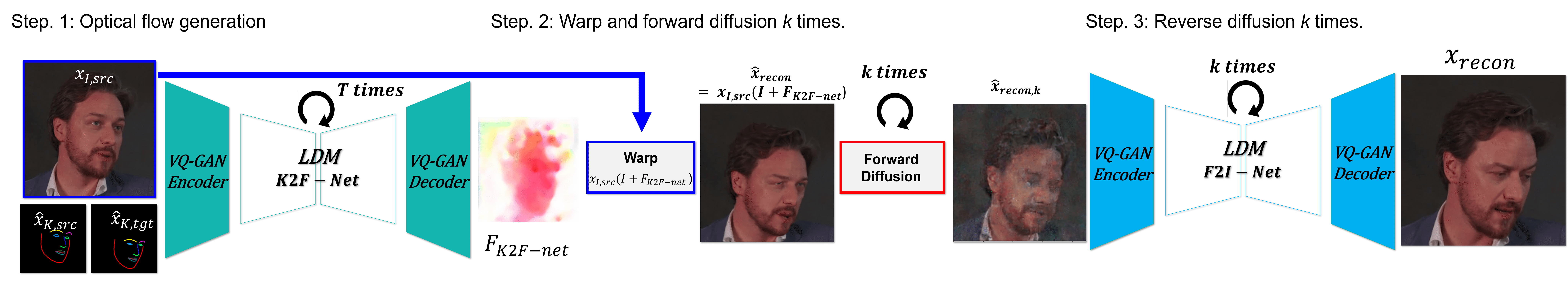}
\caption{Overview of the image manipulation process.}
\label{fig2}
\end{figure}
\noindent Figure~\ref{fig2} illustrates the image generation process employed to enhance the stability of the KDM framework. By linearly deforming the source image $x_{\text{I,src}}$ using the optical flow field $F_{\text{F2I-net}}$, provided by the F2I-net, a pseudo image reconstruction $\hat{x}_{\text{recon}} = x_{\text{I,src}} (I+ F_{\text{F2I-net}})$ is obtained. We observed that $\hat{x}_{\text{recon}}$ serves as an improved initialization condition for $x_{\text{recon}}$ generation of the F2I-net. Instead of initiating denoising from $\mathcal{N}(0,1)$, we propose leveraging the noisy pseudo-image $\hat{x}_{\text{recon},k}$, generated through $k$ iterations of forward diffusion of $\hat{x}_{\text{recon},k}$. Starting from $\hat{x}_{\text{recon},k}$, a synthesis of $x_{\text{recon}}$ is achieved through $k$ iterations of the reverse diffusion process using the F2I-net. The proposed inference procedure contributes to maintaining sequential consistency of the image background and hue, which is well-regularized within $\hat{x}_{\text{recon}}$.

\section{Experiments}
\label{section41}
In this section, the evaluation of the proposed KDM framework is demonstrated. The framework is applied to two major computer vision tasks, key-point-based human pose synthesis, and landmark guided facial image synthesis, along with a medical image generation task focused on echocardiography video synthesis. We begin by introducing details of the dataset employed for each task, then provide qualitative and quantitative results.

\subsection{Experiments setup}
\textbf{Dataset.} For facial expression image synthesis, CelebV-HQ Dataset~\cite{langner2010presentation}, a large scale facial attributes video dataset is employed. The \textit{dlib}~\cite{king2009dlib} library is adopted to identify 68 facial landmark points $x_K$. The points are reconfigured into 11 distinct colored lines $\hat{x}_K$, which include the lower face contour, left eyebrow, right eyebrow, left eye, right eye, nose bridge, nose bottom line, and four lip lines. The dataset was split into subsets of 33,000 and 3,000 for training and testing, respectively.

For human pose generation, IXMAS action dataset~\cite{weinland2006free}, which includes 11 actions performed by 10 subjects from 5 camera viewpoints, is used. The \textit{OpenPose}~\cite{cao2017realtime} was utilized to extract 25 key-points indicating body joints $x_K$. $x_K$ were interconnected with colored lines along the anatomical structure of the human body, forming a skeletal framework $\hat{x}_K$. The dataset comprises a total of 1148 sequences, with 1048 used for training and 100 for testing.

In medical imaging, EchoNet-Dynamic dataset~\cite{ouyang2020video}, which contains a collection of apical 4-chamber (A4C) echocardiogram videos is employed. The dataset comprises 10,030 videos of individual patients. For key-point detection, 20 equidistant lines perpendicular to the vector connecting the apex and center of the basal point of the heart are formulated. The key-points are configured at the intersection of each line and the boundary of the blood pool segment $\hat{x}_K$, resulting in a total of 40 key-points $x_K$. 

All datasets were resized to $224 \times 224 \times 3$ pixels.\\

\noindent\textbf{Implementation details.} AdamW~\cite{loshchilov2017decoupled} optimizer is employed for the training of the KDM. In order to enhance computational efficiency during the sampling procedure, the images are generated under DDIM process with $T=200$ denoising steps.\\

\noindent\textbf{Evaluation metrics.} The performance of generated images is assessed with diverse image and key-point-related metrics. Mean Squared Error ($MSE_{\text{img}}$) was introduced to evaluate the pixel-level similarity of the images. Perceptual similarity in the images was quantified by employing \textit{LPIPS}~\cite{zhang2018unreasonable}. The  $\textit{LPIPS}_{\text{alx}}$ and $\textit{LPIPS}_{\text{vgg}}$ employ the AlexNet~\cite{krizhevsky2012imagenet} and VGG16~\cite{simonyan2014very}, pre-trained with the ImageNet~\cite{deng2009large} dataset, as baseline structures. In addition, in order to assess whether the generated image properly integrates the key-point condition, we identified landmarks of the synthesized images and evaluated the mean distance (\textit{MD}) with the ground-truth landmarks.\\

\noindent\textbf{Comparative studies.} Comparative studies are performed by assessing the KDM framework with drag-based editing models, including UserControllableLT~\cite{endo2022user}, DragGan~\cite{pan2023drag}, and DragDiffusion~\cite{shi2023dragdiffusion}, as well as key-point guided image translation models such as PG2~\cite{ma2017pose}, C2GAN~\cite{tang2019cycle}, and PIDM~\cite{bhunia2023person}.

For the comparative study, the single-point editing process of UserControllableLT is extended to multiple key-point manipulations by formulating flow vectors of each key-point. DragGan enables drag-based multi-point manipulation through iterative optimization of the latent space of a pre-trained generative adversarial network. In UserControllableLT and DragGan, the PTI inversion~\cite{roich2022pivotal} is implemented to embed an image into the latent space. DragDiffusion achieves drag-based diffusion probabilistic image synthesis by employing the LoRA scheme~\cite{hu2021lora}. For validation of drag-based editing models, we configure the number and value of the editing vectors to that of key-points. PG2 proposes to generate person images using target key-points and input image employing a dual-generator model. C2GAN enhances robustness of the key-point conditioned image synthesis through end-to-end training of three cycles generative models. PIDM achieves key-point guided image generation by employing DDPM with cross-attentional conditioning. To ensure a fair evaluation, each model is fine-tuned with the corresponding dataset.

\section{Face Image Synthesis}
\begin{figure}[h]
\centering
\includegraphics[width=\textwidth]{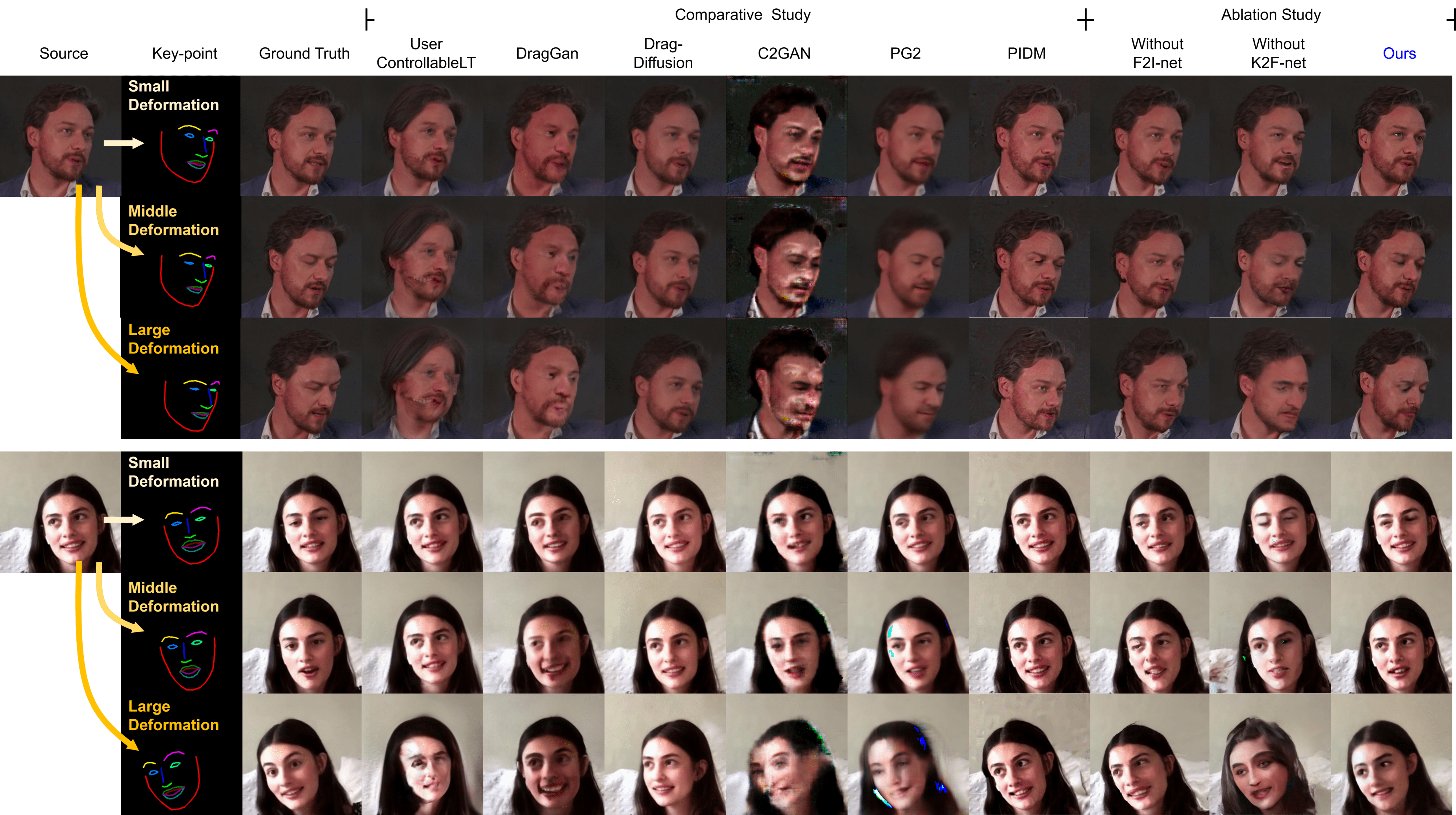}
\caption{Qualitative studies of the facial landmark guided image synthesis.}
\label{fig3}
\end{figure}
\noindent\textbf{Qualitative evaluation.} Figure~\ref{fig3} shows the qualitative comparison between the proposed model, and K2F-net and F2I-net ablated baselines. A facial image and corresponding facial landmark are given as source input. Three sets of facial landmarks, featuring from small to large deformations, are presented as target key-point conditions. The proposed framework successfully generates the facial image without compromising the style feature of the original image. In the case of an F2I-net ablated baseline, global features such as eye shape, are effectively generated. However, the network demonstrates limitations in expressing detailed image features, since the model relies on linear spatial domain transformation provided by the optical flow estimates. In the case of the K2F-net ablated baseline, image generation is solely regulated by sparse key-point conditions, resulting in each facial image exhibiting relatively inconsistent global style features when compared to the proposed framework.

The qualitative analysis of the comparative studies is introduced in Figure~\ref{fig3}. Overall, the key-point-based synthesizing model outperforms the drag-based image manipulation model in providing a faithful image. The DragGan and DragDiffusion have limited effectiveness in describing sophisticated facial expressions compared to key-point-based schemes. The C2GAN and PG2 demonstrate an enhanced understanding of the semantic relation between facial key-points and input images. However, the completeness of the image is lacking compared to the proposed model, since generating realistic images remains challenging for one-shot generative adversarial networks. The PIDM achieves realistic image generation through DDPM, however, it demonstrates challenges in preserving the attributes of the original image. On the other hand, the proposed KDM shows excellence in accurately capturing the features of the input image.\\
\begin{table}[h]
\centering
\caption{Quantitative study of the KDM framework on facial expression synthesis.}
\label{table1}
\resizebox{0.8\textwidth}{!}{%
\begin{tabular}{cc|cccc}
\hline
\multicolumn{2}{c|}{Model}             & $\textit{MSE}_{\text{img}}$ & \textit{MD}      & $\textit{LPIPS}_{\text{alx}}$ & $\textit{LPIPS}_{\text{vgg}}$ \\ \hline
\multicolumn{2}{c|}{No-edit}           & 0.0312   & 19.87 & 0.2707     & 0.3341     \\ \hline
\multicolumn{1}{c|}{\multirow{3}{*}{Drag-based}}        & UserControllableLT & 0.0322 & 16.73 & 0.3098 & 0.4079 \\
\multicolumn{1}{c|}{} & DragGan        & 0.0277   & 13.72 & 0.2919     & 0.3901     \\
\multicolumn{1}{c|}{} & DragDiffusion & 0.0219   & 4.787  & 0.2029     & 0.2735     \\ \hline
\multicolumn{1}{c|}{\multirow{3}{*}{Key-point guided}} & C2GAN              & 0.0214 & 4.194  & 0.3084 & 0.3908 \\
\multicolumn{1}{c|}{} & PG2            & \textbf{0.0177}   & 2.397  & 0.2909     & 0.3162     \\
\multicolumn{1}{c|}{} & PIDM           & 0.0198   & 2.944   & 0.2164     & 0.3166     \\ \hline
\multicolumn{1}{c|}{\multirow{3}{*}{Proposed}}         & w.o. F2I-net       & 0.0188 & 3.911   & 0.1880  & 0.2767 \\
\multicolumn{1}{c|}{} & w.o. K2F-net   & 0.0222   & 2.773   & 0.2333     & 0.3161     \\
\multicolumn{1}{c|}{} & \textbf{Ours}           & \textbf{0.0180}   & \textbf{1.874}   & \textbf{0.1739}     & \textbf{0.2572}     \\ \hline
\end{tabular}%
}
\end{table}

\noindent\textbf{Quantitative evaluation.} The quantitative evaluation of facial image synthesis is introduced in Table~\ref{table1}. The drag-based image manipulation of the UserControllableLT causes undesired change of the source image, resulting in the Mean Squared Error $MSE_{\text{img}}$ being worse than the original image with no modification. The DragGan and DragDiffusion demonstrate enhanced image manipulation compared to UserControllableLT, however showing lower \textit{MD} compared to the key-point guided generative model, proving the key-point guided supervision offers a precise semantic understanding of facial landmarks. However, the key-point guided generative adversarial networks, which are C2GAN and PG2 show 0.134 and 0.088 worse \textit{LPIPS} distance than the proposed scheme, indicating the GAN-based model has limited efficacy in producing an authentic face reconstruction.

In the ablation study, the F2I-net ablated model demonstrates acceptable performance in $\text{MSE}_{\text{img}}$ metrics, however, shows limitations when it is evaluated with perceptual scores. This is due to the aforementioned fact that linear deformation cannot adequately describe high-frequency details. In overall evaluation metrics, the proposed optical flow-assisted framework outperforms the K2F-net ablated one, proving the key-point to the dense optical flow completion process is essential for high-quality image generation.\\

\begin{figure}[h]
\centering
\includegraphics[width=\textwidth]{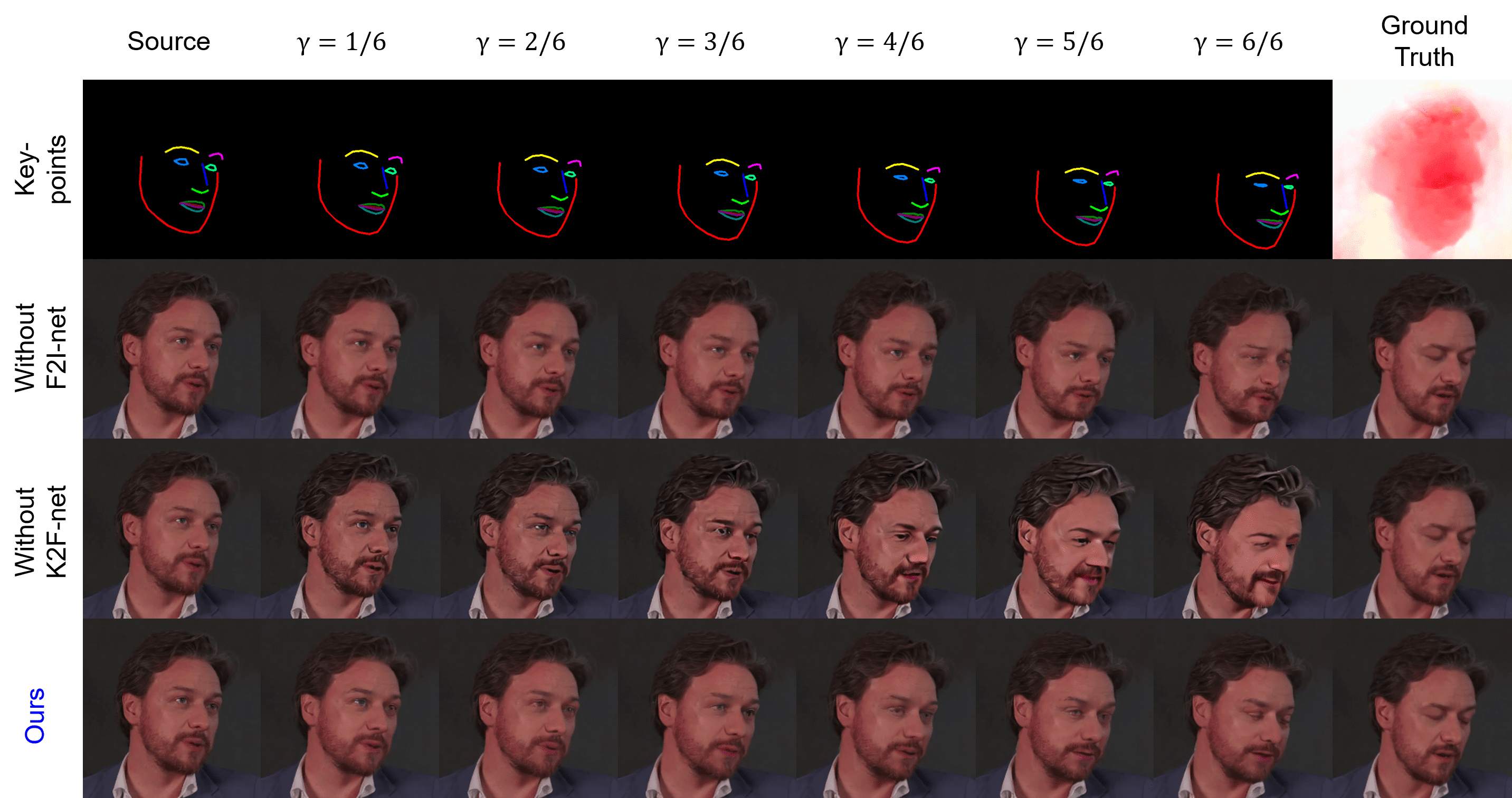}
\caption{Qualitative result on the continuous facial image generation. In order to visualize the optical flow fields, the flow field is color-coded as proposed in~\cite{butler2012naturalistic}.}
\label{fig4}
\end{figure}

\noindent\textbf{Continuous image generation.} Furthermore, we performed continuous generation of facial images sequentially transitioning from source to target expression. From a single generation of optical flow $F$, we can scale the deformation field as $\gamma \cdot F$, where $0 \leq \gamma \leq 1$. Through the F2I-net denoising process $\phi_{F2I} (x_{\text{I,src}}, \gamma \cdot F, t)$ of the scaled deformation field $\gamma \cdot F$, the intermediate frame image is generated. Figure~\ref{fig4} presents continuous image generation where $\gamma = (1/6, 2/6, 3/6, ... ,6/6)$. The neural network realistically produces successive transformations in a person's facial expression. On the other hand, the K2F-net ablated baseline, where the image generation is only conditioned by sparse key-points, struggles to maintain sequential consistency, resulting in inauthentic image generation.
\begin{figure}[t]
\centering
\includegraphics[width=0.9\textwidth]{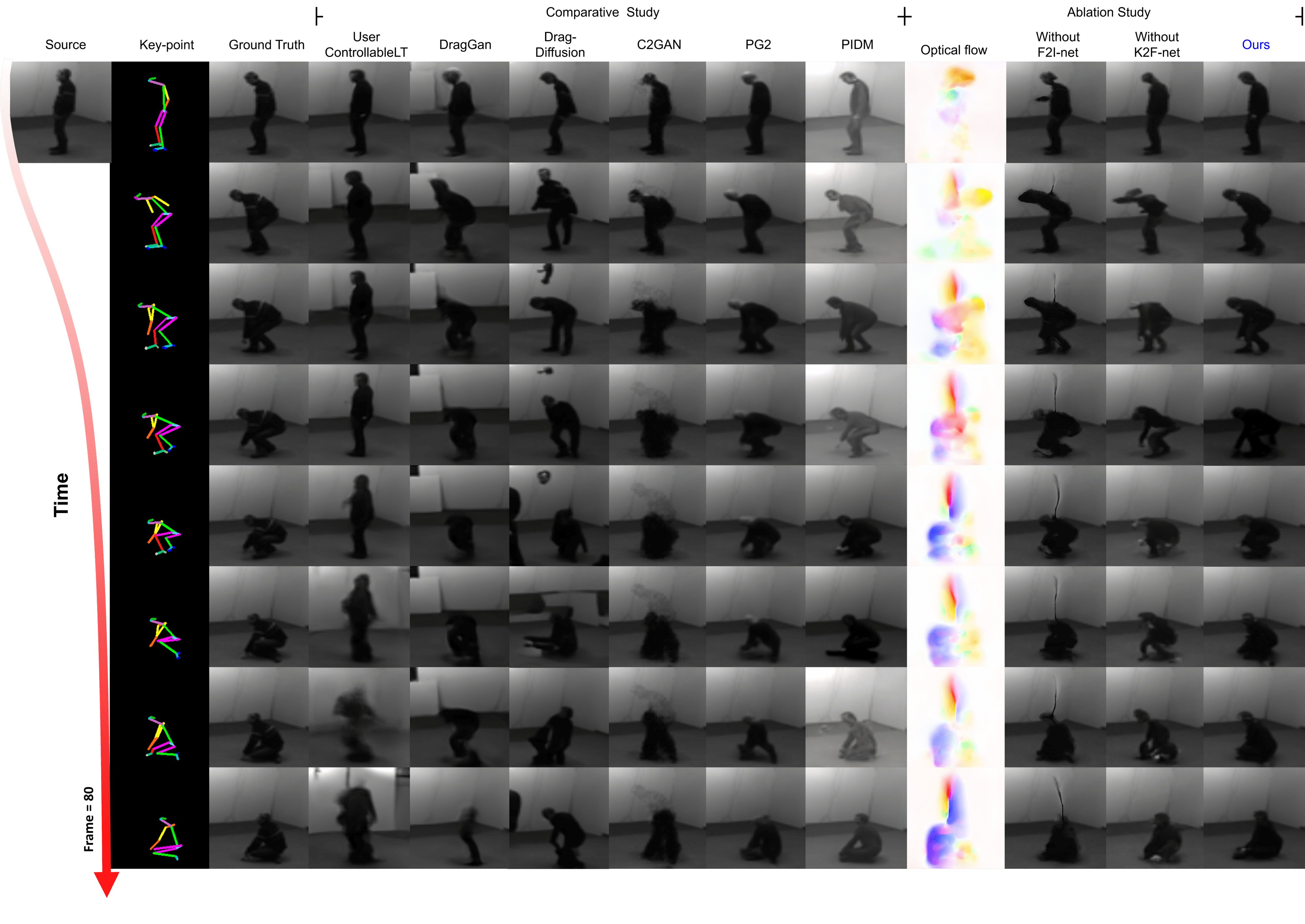}
\caption{Qualitative study of the key-point guided human pose synthesis.}
\label{fig5}
\end{figure}
\subsection{Pose synthesis}
\noindent\textbf{Qualitative evaluation.} Qualitative assessment is presented in the Figure~\ref{fig5}. The qualitative assessment shows a sequential image generation of a standing man transitioning into a sitting position. The proposed framework succeeds in generating sequential images with precision. Compared to the K2F-net ablated baseline network, the KDM framework demonstrates excellence in forming stable sequential images. When the F2I-net is ablated, the neural network fails to achieve sophisticated image synthesis. Especially, when a significant amount of key-point control is applied, the neural network often leaves warped afterimage artifacts. 
\begin{table}[t]
\centering
\caption{Quantitative study of the KDM framework on the human pose synthesis.}
\label{table2}
\resizebox{0.8\textwidth}{!}{%
\begin{tabular}{cc|cccc}
\hline
\multicolumn{2}{c|}{Model}            & $\textit{MSE}_{\text{img}}$ & \textit{MD}     & $\textit{LPIPS}_{\text{alx}}$ & $\textit{LPIPS}_{\text{vgg}}$ \\ \hline
\multicolumn{2}{c|}{No-edit}          & 0.0080   & 24.32 & 0.1773     & 0.2060     \\ \hline
\multicolumn{1}{c|}{\multirow{3}{*}{Drag-based}}        & UserControllableLT & 0.0135 & 29.57 & 0.2826 & 0.3388 \\
\multicolumn{1}{c|}{} & DragGan       & 0.0052   & 23.54 & 0.1645     & 0.2679     \\
\multicolumn{1}{c|}{} & DragDiffusion & 0.0074   & 21.92 & 0.1758     & 0.2108     \\ \hline
\multicolumn{1}{c|}{\multirow{3}{*}{Key-point guided}} & C2GAN              & 0.0017 & 16.06 & 0.0925 & 0.1731 \\
\multicolumn{1}{c|}{} & PG2           & \textbf{0.0009}   & 12.74 & 0.0711     & 0.1199     \\
\multicolumn{1}{c|}{} & PIDM          & 0.0742   & 17.07 & 0.3238     & 0.3995     \\ \hline
\multicolumn{1}{c|}{\multirow{3}{*}{Proposed}}         & w.o. F2I-net       & 0.0013 & 14.00 & 0.1068 & 0.1494 \\
\multicolumn{1}{c|}{} & w.o. K2F-net  & 0.0038   & 13.96 & 0.1107     & 0.1673     \\
\multicolumn{1}{c|}{} & \textbf{Ours}          & \textbf{0.0012}   & \textbf{10.32} & \textbf{0.0653}     & \textbf{0.1356}     \\ \hline
\end{tabular}%
}
\end{table}
In the comparative study, the effectiveness of the suggested framework is emphasized. The UserControllableLT and DragGan struggle to isolate the background from the objects. Consequently, background artifacts occur in such drag-based generative adversarial networks, resulting in undesirable sequential image generation. The drag-based diffusion model achieves enhanced image quality compared to DragGan, however, demonstrates limitations in capturing sophisticated key-point control. The key-point conditioned generative models are also investigated. The C2GAN and PG2 have difficulties in establishing a robust high-level semantic relation between pose change and the input image, Consequently, they struggle to accurately generate detailed images, particularly when faced with significant key-point deformations. For example, both of the neural networks fail to generate the upper body part of the images. On the other hand, the DDPM-based models, which are PIDM and KDM, achieve reliable image generation, precisely illustrating every body part of the human. The PIDM demonstrates limited consistency in consecutive image generation, while the proposed two-stage KDM framework succeeds in providing faithful sequential human pose generation.\\

\noindent\textbf{Quantitative evaluation.} Table~\ref{table2} shows quantitative assessments of the KDM scheme and baseline models. The proposed scheme outperforms the F2I-net and K2F-net ablated baselines for every metric. The F2I-net ablated baseline demonstrates good performance in $\textit{MSE}_{\text{img}}$, since the $\textit{MSE}_{\text{img}}$ metric largely depends on the global feature of the image and such low-frequency image features can be generated through the linear deformation with optical flow displacement. However, the network demonstrates limited performance when it is evaluated with perceptual metrics that assess the authenticity of the generated image. In the comparative study, the proposed neural network outperforms drag-based generative models by a considerable margin. In the case of key-point conditioned generative adversarial networks, the PG2 demonstrates comparable performance in $\textit{MSE}_{\text{img}}$. However, the \textit{MD} and \textit{LPIPS} show that the proposed two-stage generative scheme demonstrates excellence in interpreting the semantic relation between key-point and source image, thus yielding desirable image outcome.
\begin{figure}[h]
\centering
\includegraphics[width=\textwidth]{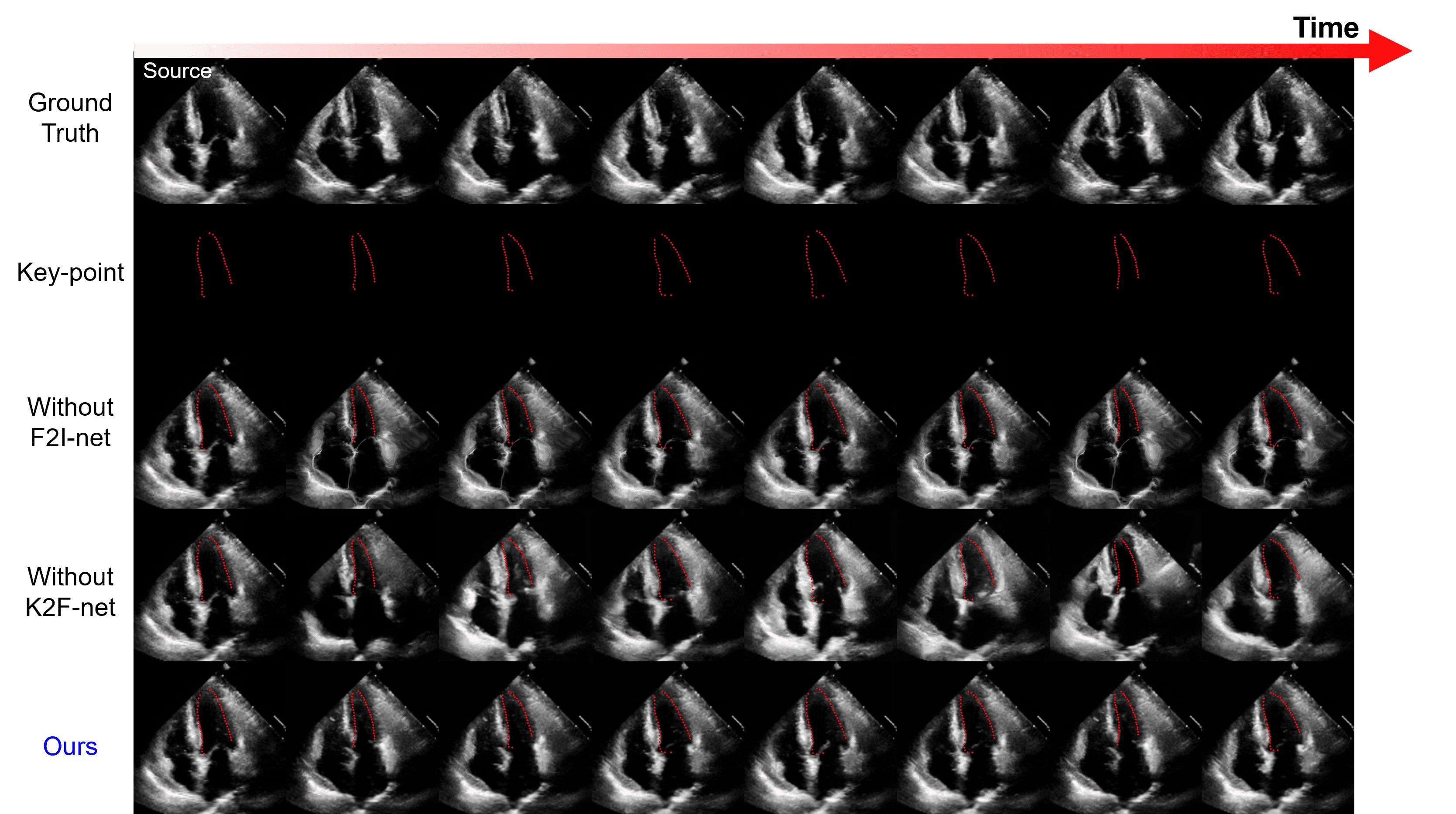}
\caption{Qualitative study of the echocardiography video synthesis.}
\label{fig6}
\end{figure}
\begin{table}[h]
\centering
\caption{Quantitative study of the KDM framework on the echocardiography video synthesis.}
\label{table3}
\begin{adjustbox}{width=0.5\textwidth}
\begin{tabular}{c|ccc}
\hline
Model        &$\textit{MSE}_{\text{img}}$ & $\textit{LPIPS}_{\text{alx}}$ &$\textit{LPIPS}_{\text{vgg}}$ \\ \hline
w.o. F2I-net & \textbf{0.0121}   & 0.1980      & 0.3028     \\
w.o. K2F-net & 0.0268   & 0.2375     & 0.3119     \\
\textbf{Ours}         & \textbf{0.0124}   & \textbf{0.1739}     & \textbf{0.2645}     \\ \hline
\end{tabular}
\end{adjustbox}
\end{table}
\subsection{Echocardiography video synthesis}
An echocardiography video synthesis task of generating time sequential cardiac movement from a single cardiac image is investigated. In echocardiography, high-frequency details, such as acoustic speckles, are clinically important observations. Therefore, fine-grained image synthesis is important. Figure~\ref{fig6} presents generated echocardiography images of the KDM and the baselines. The KDM achieves detail-preserved image synthesis which was not possible in the F2I-net ablated one. The F2I-net ablated baseline demonstrates comparable $\textit{MSE}_{\text{img}}$ with the proposed model, however, this is attributed to the fact that the optical flow is optimized to minimize the mean squared difference between the ground-truth and the warped image. However, the proposed method demonstrates significantly enhanced performance when it is compared with \textit{LPIPS} metrics. Furthermore, the KDM framework demonstrates excellence in style-consistent image generation under continuous manipulation of cardiac key-points, proving the proposed scheme provides realistic echocardiography video synthesis.

\section{Conclusion}
In this paper, we propose a key-point guided diffusion probabilistic model (KDM), a cross-modal framework that provides interactive key-point guided image generation. A stochastic key-point to optical flow generative model is implemented, constructing a dense optical flow matrix, and interpreting high-level semantic relation between the input image and the key-point control. The performance of the proposed framework is investigated in facial expression generation, human pose synthesis, and echocardiography video generation tasks. Results across various image synthesis datasets demonstrate that the KDM provides fine-grained image generation compared to baseline networks while showing enhanced sequential consistency attributed to the utilization of the optical flow.

% \section{Acknowledgement}
% This work was supported by the anonymous project grant funded by the anonymous organization (the anonymous institute A, the anonymous institute B, the anonymous institute C, the anonymous institute D, and the anonymous institute E).

\clearpage

\appendix
\section*{Appendix}

\noindent The supplementary materials demonstrate the evaluation of the optical flow estimation (Section~\ref{Sup_section1}), effectiveness of the proposed continuous image generation (Section~\ref{Sup_section2}), examination of the impact of initialization with linear deformation (Section~\ref{Sup_section3}),include additional results on drag-based image editing (Section~\ref{Sup_section4}), and discussion about limitation (Section~\ref{Sup_section5}).

\section{Optical Flow Estimation}
\label{Sup_section1}
The following sections include assessments and discussions of the optical flow estimation schemes employed in the KDM framework. Section~\ref{Sup_section11} demonstrates the accuracy of the optical flow estimation network (OF-net). Section~\ref{Sup_section12} presents the effectiveness of the key point guided flow synthesis network (K2F-net) in generating the optical flow estimates.

\subsection{Assessment of the OF-net.}
\label{Sup_section11}
\begin{figure}[h]
\centering
\includegraphics[width=\textwidth]{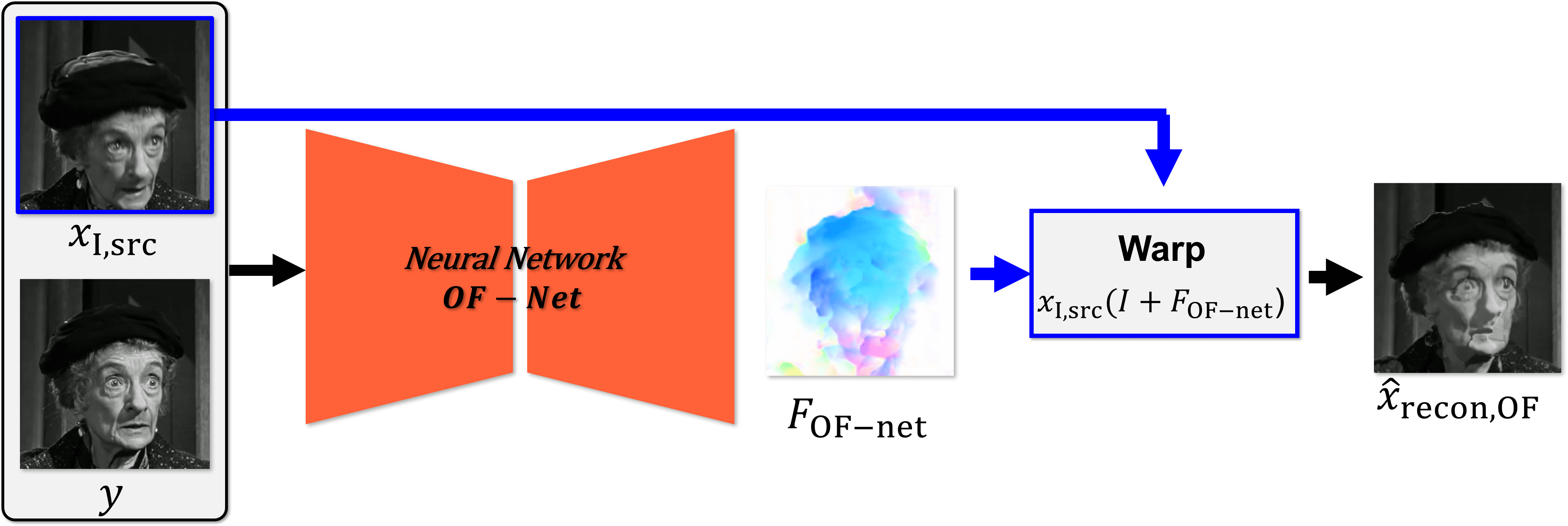}
\caption{Illustration of the $\hat{x}_{\text{recon,OF}}$ generation procedure.}
\label{Sup_fig1}
\end{figure}

\begin{table}[h]
\centering
\caption{Quantitative assessment of the OF-net and K2F-net.}
\label{Sup_table1}
\begin{adjustbox}{width=0.7\textwidth}
\begin{tabular}{cc|ccc}
\hline
\multicolumn{2}{c|}{Model}                                       & $MSE_{img}$ & $LPIPS_{alx}$ & $LPIPS_{vgg}$ \\ \hline
\multicolumn{1}{c|}{\multirow{2}{*}{Facial image}}     & K2F-net & 0.0222   & 0.2333     & 0.3161     \\
\multicolumn{1}{c|}{}                                  & OF-net  & 0.0080   & 0.1083     & 0.1709     \\ \hline
\multicolumn{1}{c|}{\multirow{2}{*}{Human pose}}       & K2F-net & 0.0013   & 0.1068     & 0.1494     \\
\multicolumn{1}{c|}{}                                  & OF-net  & 0.0002   & 0.0527     & 0.0831     \\ \hline
\multicolumn{1}{c|}{\multirow{2}{*}{Echocardiography}} & K2F-net & 0.0121   & 0.1980     & 0.3028     \\
\multicolumn{1}{c|}{}                                  & OF-net  & 0.0014   & 0.1128     & 0.2231     \\ \hline
\end{tabular}
\end{adjustbox}
\end{table}

\begin{figure}[h]
\centering
\includegraphics[width=0.98\textwidth]{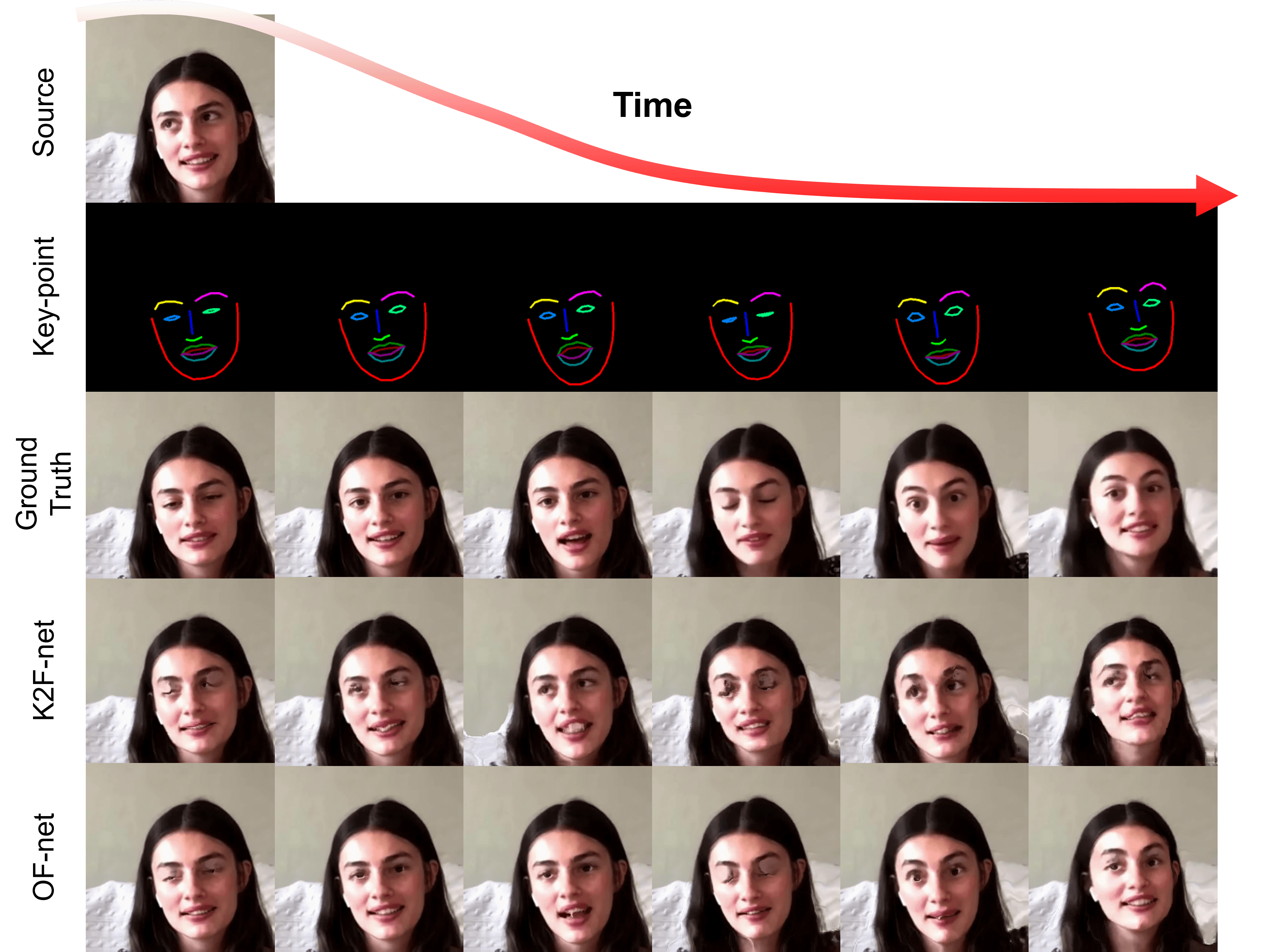}
\caption{Qualitative assessment of the K2F-net and OF-net for facial image generation.}
\label{Sup_fig2}
\end{figure}

\begin{figure}[!htb]
  \centering
  \includegraphics[width=0.98\textwidth]{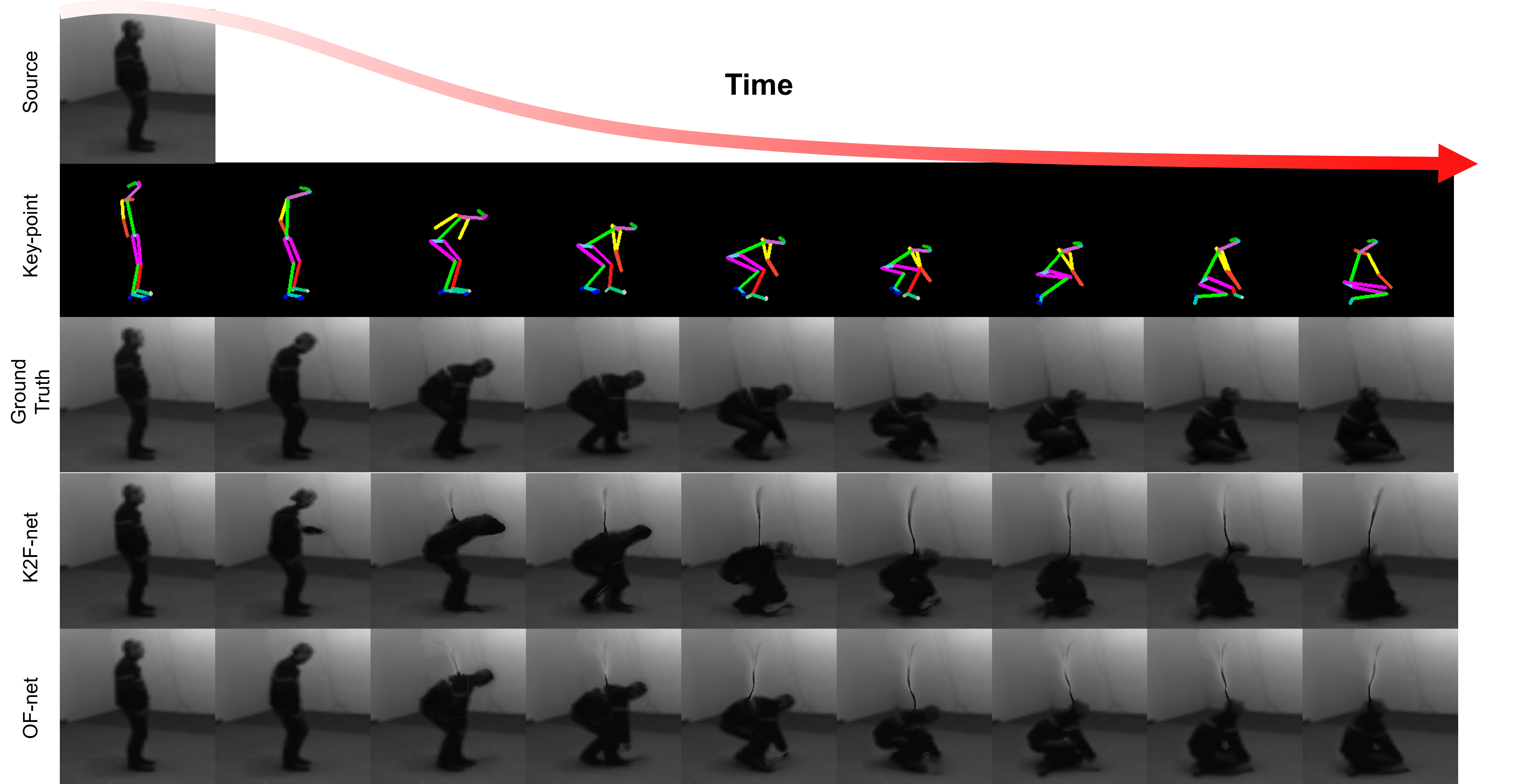}
  \caption{Qualitative assessment of the K2F-net and OF-net for human pose generation.}
  \label{Sup_fig3}
  % \vspace{-1em}
\end{figure}

\begin{figure}[!htb]
  \centering
  \includegraphics[width=0.98\textwidth]{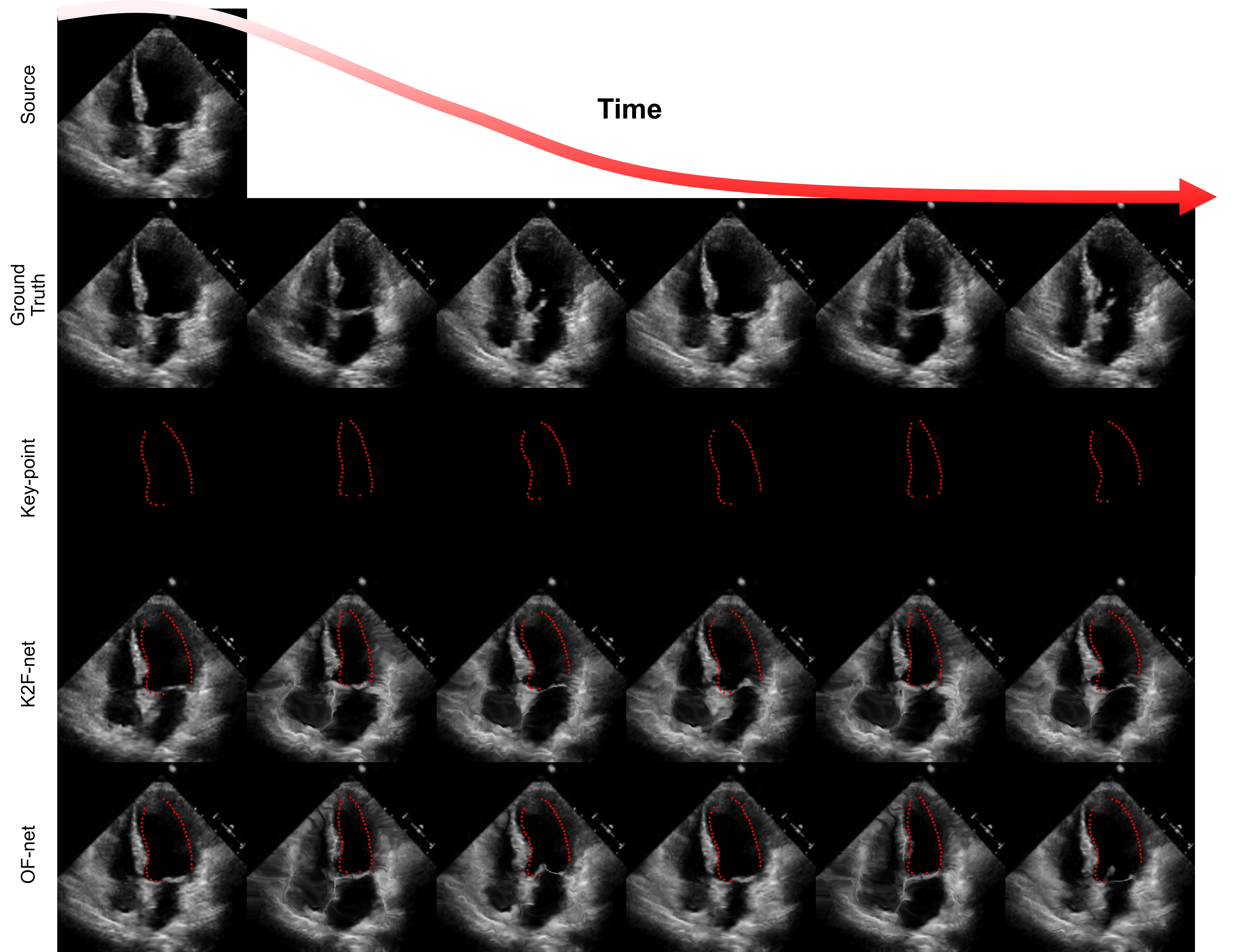}
  \caption{Qualitative assessment of the K2F-net and OF-net for echocardiography image generation.}
  \label{Sup_fig4}
\end{figure}

\noindent The OF-net generates the optical flow field \( F_{\text{OF-net}} \), employing the source \( x_{\text{I,src}} \) and target \( y \) images as inputs. The warped target image \( x_{\text{recon,OF}} \) is generated through linear deformation of the \( x_{\text{I,src}} \) with \( F_{\text{OF-net}} \) (described in Figure~\ref{Sup_fig1}). The quantitative assessment is performed by comparing the warped image \( \hat{x}_{\text{recon,OF}} \) with the ground-truth target image \( y \). The Mean Squared Error ($MSE_{img}$) and \textit{LPIPS} metrics, described in the main script, are employed as evaluation metrics. The OF-net demonstrates \( 0.0080 \) and \( 0.0002 \) $MSE_{img}$ for the facial image and human pose datasets, respectively. Figure~\ref{Sup_fig2},~\ref{Sup_fig3}, and~\ref{Sup_fig4} present a qualitative assessment of the \( \hat{x}_{\text{recon,OF}} \) for the facial image, human pose, and echocardiography generation, respectively. The global features of the \( \hat{x}_{\text{recon,OF}} \), such as the shape of the mouth, are accurately generated.

\subsection{Assessment of the K2F-net.}
\label{Sup_section12}
\begin{table}[!htb]
\centering
\caption{Quantitative assessment of the K2F-net in optical flow estimation.}
\label{Sup_table2}
\begin{adjustbox}{width=0.5\textwidth}
\begin{tabular}{cc|cc}
\hline
\multicolumn{2}{c|}{Model}                                                           & $MAE$ & $RMSE$ \\ \hline
\multicolumn{1}{c|}{\multirow{3}{*}{K2F-net}} & \begin{tabular}[c]{@{}c@{}}Facial image\end{tabular} & 6.936 & 9.720 \\
\multicolumn{1}{c|}{} & \begin{tabular}[c]{@{}c@{}}Human pose\end{tabular}         & 3.535 & 7.032  \\
\multicolumn{1}{c|}{} & \begin{tabular}[c]{@{}c@{}}Echocardiography\end{tabular} & 2.231 & 3.471  \\ \hline
\end{tabular}
\end{adjustbox}
\end{table}

The K2F-net generates the optical flow \( F_{\text{K2F-net}} \), employing the key-point \( x_{\text{K}} \) and \( x_{\text{I,src}} \). The performance of the K2F-net is evaluated by comparing the accuracy of the \( F_{\text{K2F-net}} \) with the \( F_{\text{OF-net}} \). The Root Mean Squared Error (\textit{RMSE}) and Mean Absolute Error (\textit{MAE}) are employed as the evaluation metrics. Table~\ref{Sup_table2} demonstrates quantitative assessments of the K2F-net. The K2F-net demonstrates 9.720 \textit{RMSE} for facial optical flow estimation. Figure~\ref{Sup_fig2}-\ref{Sup_fig4} 3rd rows present a warped image \( x_{\text{I,src}} (I+F_{\text{K2F-net}}) \) using \( F_{\text{K2F-net}} \). The evaluation demonstrates that the K2F-net achieves precise estimation of the optical flow using the key-point condition.

\begin{figure}[htb!]
\centering
\includegraphics[width=1\textwidth]{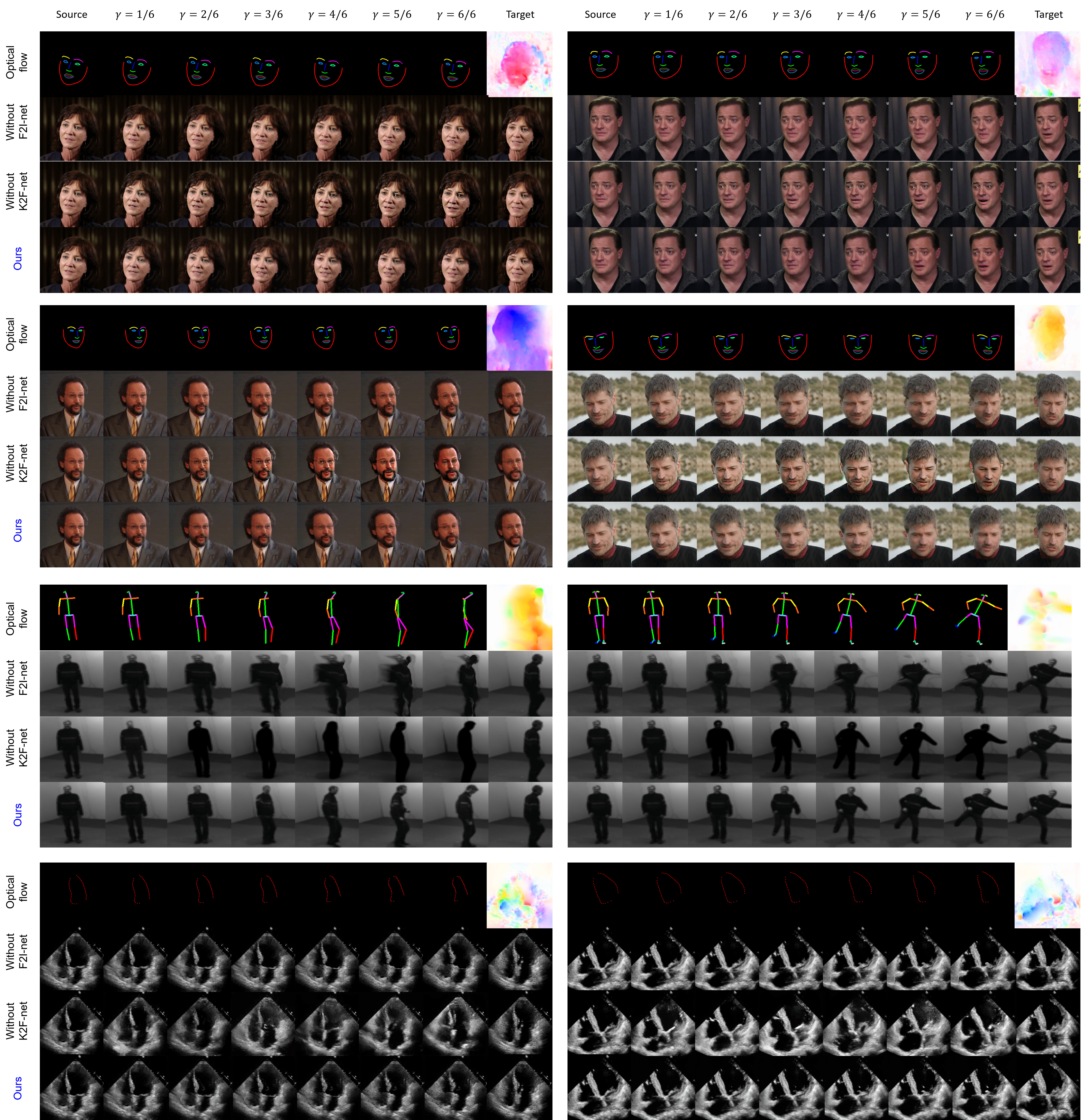}
\caption{Qualitative assessment of continuous image generation of facial image synthesis, human pose generation and echocardiography video prediction.}
\label{Sup_fig5}
\end{figure}

\section{Continuous Image Generation}
\label{Sup_section2}
In this section, additional experiments on continuous image generation are provided. The continuous sequential images are synthesized by the following procedure. Firstly, an optical flow matrix \( F_{\text{K2F-net}} \) is generated through the denoising process of the K2F-net. Subsequently, the magnitude of \( F_{\text{K2F-net}} \) is adjusted to \( \gamma \cdot F_{\text{K2F-net}} \), where \( 0 \leq \gamma \leq 1 \). The \( \gamma \cdot F_{\text{K2F-net}} \) is employed as an optical flow of the intermediate frame image. Finally, the F2I-net generates intermediate frame images \( \phi_{\text{F2I}}(x_{\text{I,src}}, x_{\text{K}}, \gamma \cdot F, t) \), utilizing the intermediate optical flow \( \gamma \cdot F_{\text{K2F-net}} \) and the source image \( x_{\text{I,src}} \). The proposed continuous image generation scheme enhances the efficiency of the KDM framework by reducing the need to infer optical flow for every intermediate frame image. Figure~\ref{Sup_fig5} presents continuous image generation of the human face, pose, and echocardiography images, where \( \gamma = (\frac{1}{6}, \frac{2}{6}, \frac{3}{6}, \ldots, \frac{6}{6}) \). The KDM framework succeeds in generating sequentially consistent and authentic images. The continuous generation of KDM can be applied to a range of applications such as video generation and image frame interpolation.

\section{Ablation Study: Initialization with Linear Deformation}
\label{Sup_section3}
\begin{figure}[h]
\centering
\includegraphics[width=0.98\textwidth]{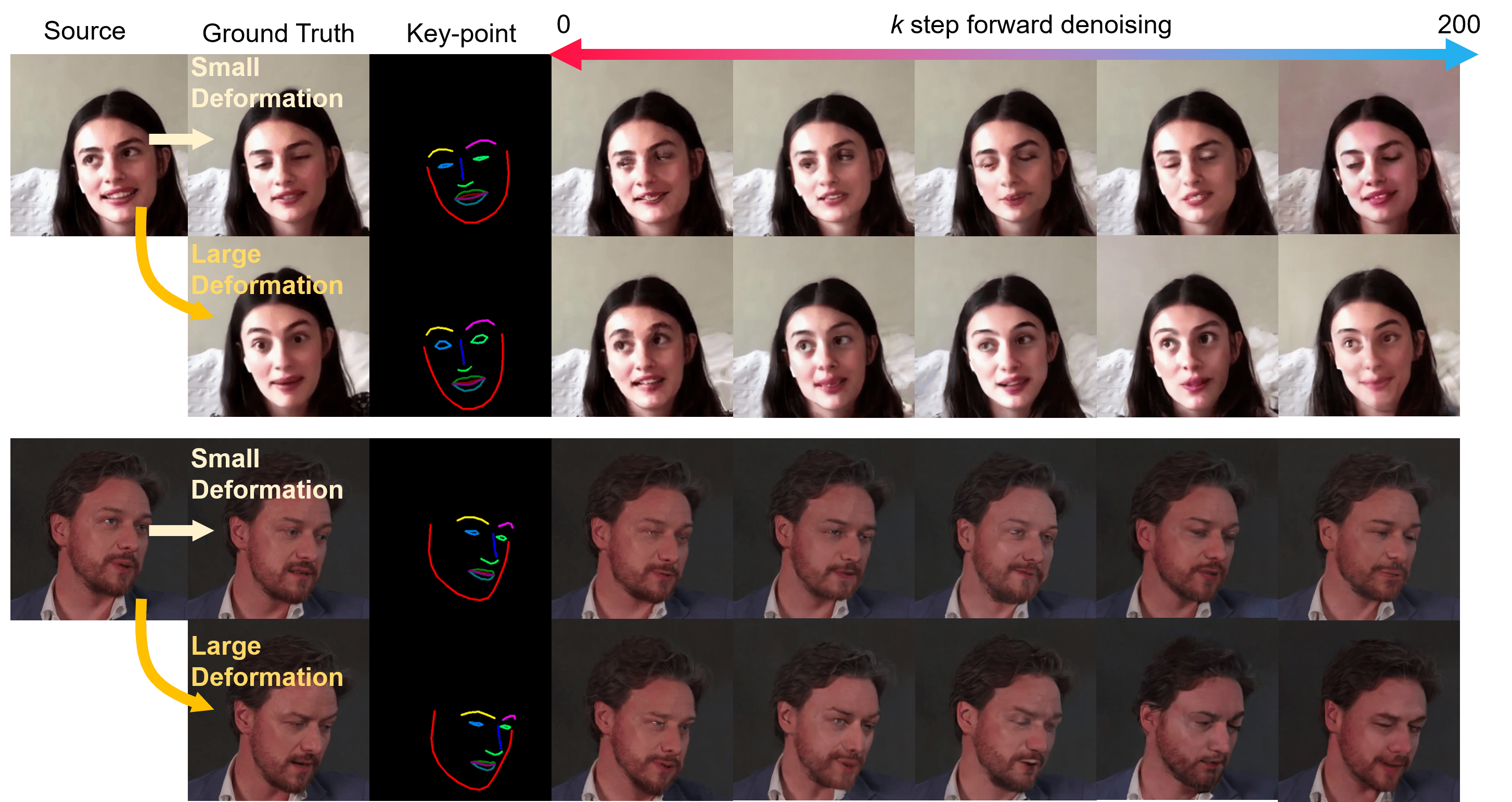}
\caption{Qualitative assessment of the image initialization with varying $k$ variable is provided.}
\label{Sup_fig6}
\end{figure}

\noindent In this section, we conduct ablation studies on image initialization, which is described in Section 3.4 of the main paper. Figure~\ref{Sup_fig6} presents a qualitative assessment of facial image manipulation with varying \( k \) parameters. When \( k = 0 \), the F2I-net is not applied for the \( x_{\text{recon}} \) generation, and the \( x_{\text{recon}} \) is identical to \( x_{\text{I,src}} (I+ F_{\text{F2I-net}}) \). Consequently, the network fails to illustrate detailed facial expressions. For \( k = 200 \), the F2I-net initiates denoising from \( N(0,1) \), which results in limited sequential consistency of the image. To enhance sequential consistency while achieving precise image synthesis, we set \( k \) to 100.

\section{Drag-based Image Editing}
\label{Sup_section4}
\begin{figure}[h]
\centering
\includegraphics[width=0.98\textwidth]{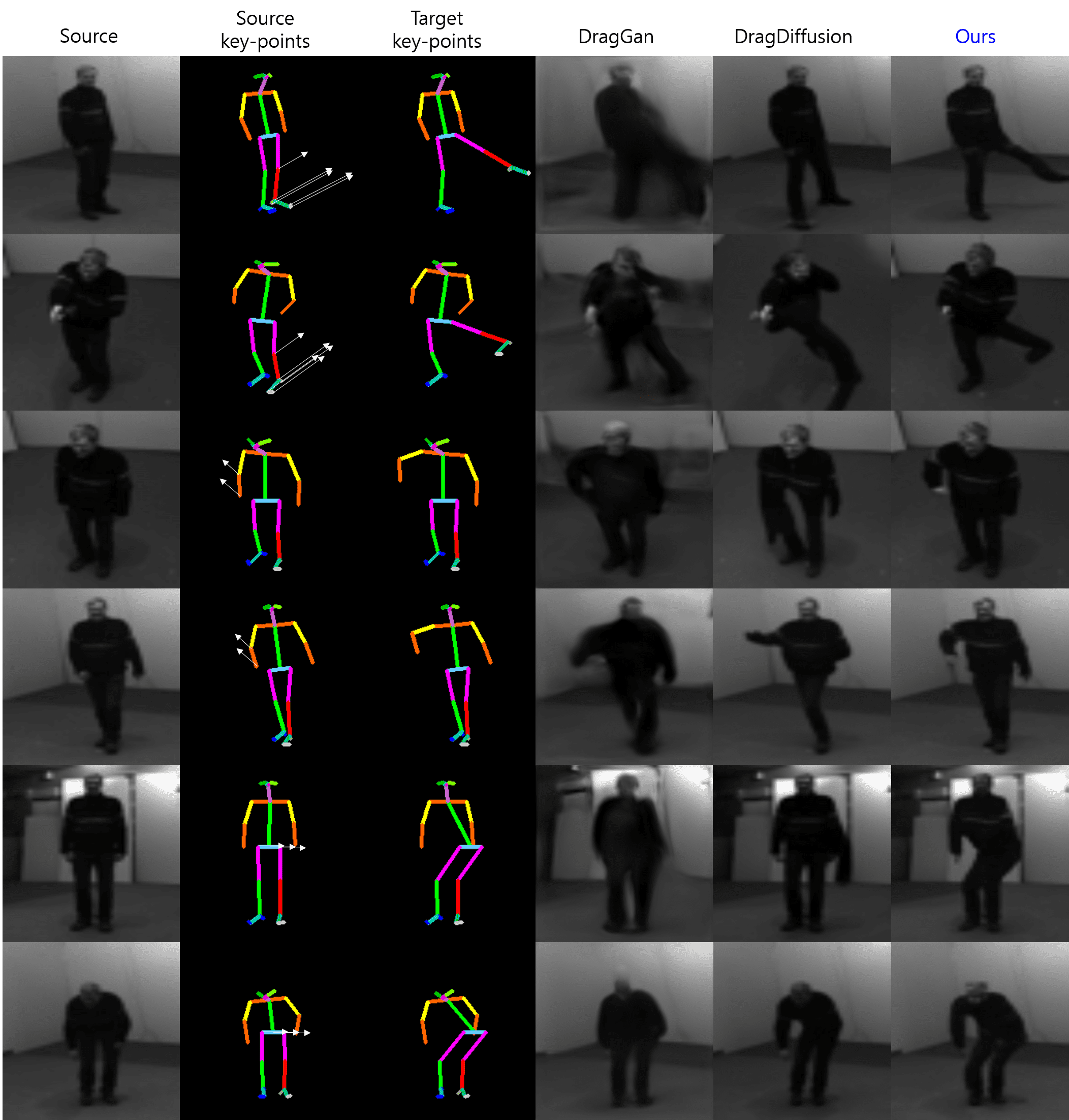}
\caption{Qualitative assessment of the drag-based image editing. White arrow in source key-points denotes drag vector \( \mathbf{v} \).}
\label{fig7}
\end{figure}
\noindent The KDM framework can be employed for drag-based image manipulation. In drag-based image editing, a user defines a motion vector \( \mathbf{v} \) through mouse dragging~\cite{endo2022user}. Then, the system provides a desired image that corresponds to the user’s drag objective. The KDM framework implements the drag-based image editing by generating the target key-point \( \mathbf{x}_{\text{K,tgt}} = \mathbf{x}_{\text{K,src}} + \mathbf{v} \), which is derived through applying motion vector \( \mathbf{v} \) to the source key-point \(\mathbf{x}_{\text{K,src}}\). Figure~\ref{Sup_fig7} presents the drag-based human pose editing of the DragGan~\cite{pan2023drag}, DragDiffusion~\cite{shi2023dragdiffusion}, and KDM framework. The KDM demonstrates versatile image generation under diverse \( \mathbf{v} \) conditions. In Figure~\ref{Sup_fig7} first row, a drag-based image editing of lifting the left leg is introduced. While DragGan and DragDiffusion show difficulties in accurately representing the modified left leg pose, the KDM achieves precise generation of every body part of the human.

\section{Limitation}
\label{Sup_section5}
\begin{figure}[h]
\centering
\includegraphics[width=0.98\textwidth]{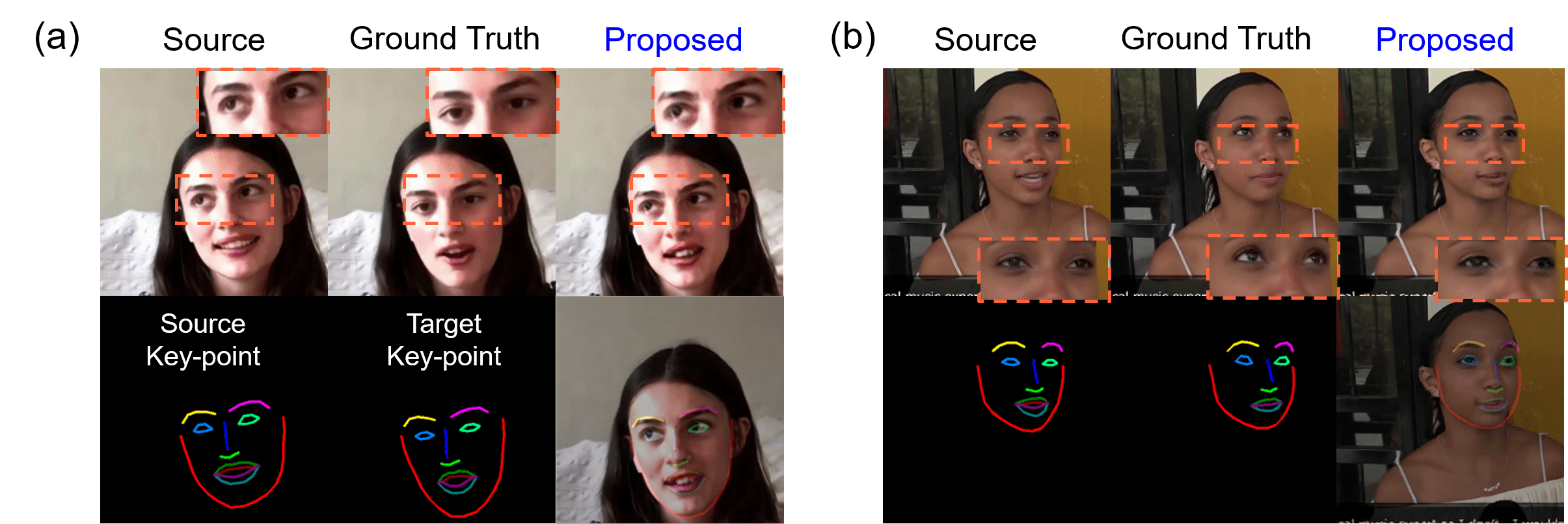}
\caption{Unnatural image examples of the KDM framework.}
\label{Sup_fig8}
\end{figure}
\noindent In this section, we present the limitations of the KDM framework. Within the KDM framework, image manipulation is restricted to objects that are semantically related to the key-points. Consequently, the versatility and effectiveness of image manipulation are significantly dependent on the configuration of the key-points. For example, in facial image manipulation, the key-point that indicates the position of the human pupil is not included among the 68 facial landmarks. As a result, the position of the pupil remains unchanged, leading to unnatural image generation, as depicted in Figure~\ref{Sup_fig8}. We anticipate configuring appropriate key-points tailored to the application of the KDM framework, thereby expanding its applicability and facilitating precise image manipulation aligned with specific objectives.

\bibliographystyle{splncs04}
\bibliography{main}

\end{document}